%% file: main.tex
\documentclass[10pt,twocolumn,letterpaper]{article}

\usepackage{wacv}              

\usepackage{graphicx}
\usepackage{tabularray}
\usepackage{adjustbox} 
\usepackage{amsmath}
\usepackage{amssymb}
\usepackage[dvipsnames]{xcolor}
\usepackage{booktabs}
\usepackage{xcolor}
\usepackage{multirow}
\usepackage{dsfont}
\usepackage{pifont}
\usepackage{tabularx}
\usepackage{color, colortbl}
\usepackage[accsupp]{axessibility}  
\definecolor{Gray}{gray}{0.9} 
\definecolor{maroon}{rgb}{0.5, 0.0, 0.0}

\newcommand{\ra}[1]{\renewcommand{\arraystretch}{#1}}
\usepackage[pagebackref,breaklinks,colorlinks]{hyperref}

\usepackage[capitalize]{cleveref}
\crefname{section}{Sec.}{Secs.}
\Crefname{section}{Section}{Sections}
\Crefname{table}{Table}{Tables}
\crefname{table}{Tab.}{Tabs.}

\def\wacvPaperID{6} 
\def\confName{WACV}
\def\confYear{2024}

\graphicspath{ {./images/} }
\begin{document}

\title{Ef-QuantFace: Streamlined Face Recognition with Small Data and Low-Bit Precision}
\author{William Gazali\textsuperscript{\dag}\qquad Jocelyn Michelle Kho\textsuperscript{\dag, \maltese}\qquad Joshua Santoso\textsuperscript{\dag}\qquad Williem\textsuperscript{\dag}\\ \\
\textsuperscript{\dag} Indonesia Vision AI
\\
\textsuperscript{\maltese} Bina Nusantara University\\
{\tt\small williamgozali2001@gmail.com, jocelyn.michelle@binus.ac.id, \{janojoshua, williem.pao\}@gmail.com}
}

\maketitle

\begin{abstract}
   
    In recent years, model quantization for face recognition has gained prominence. Traditionally, compressing models involved vast datasets like the 5.8 million-image MS1M dataset as well as extensive training times, raising the question of whether such data enormity is essential. This paper addresses this by introducing an efficiency-driven approach, fine-tuning the model with just up to 14,000 images, 440 times smaller than MS1M. We demonstrate that effective quantization is achievable with a smaller dataset, presenting a new paradigm. Moreover, we incorporate an evaluation-based metric loss and achieve an outstanding 96.15\% accuracy on the IJB-C dataset, establishing a new state-of-the-art compressed model training for face recognition. The subsequent analysis delves into potential applications, emphasizing the transformative power of this approach. This paper advances model quantization by highlighting the efficiency and optimal results with small data and training time.
     
\end{abstract}

\input{sections_workshop/section1_introduction}
\input{sections_workshop/section2_related_work}
\input{sections_workshop/section3_methodology}

\input{sections_workshop/section4_experiment_details}

\input{sections_workshop/section5_results}
\input{sections_workshop/section6_conclusion}
\input{sections_workshop/section7_future_works}

{\small
\bibliographystyle{ieee_fullname}
\bibliography{egbib}
}

\end{document}

%% file: sections_workshop/section1_introduction.tex
\section{Introduction}
\label{table:intro}

Deep neural networks have demonstrated exceptional performance across various tasks, including face recognition (FR). FR is a prominent biometric technique for identity authentication, widely employed in different sectors, including industry and security, and consistently delivers outstanding performance. A recent NIST report for FRVT (Face Recognition Vendor Test) 1:1\footnote{\textit{https://pages.nist.gov/frvt/html/frvt11.html}. Accessed on July 2023.} has shown that the top-1 rank holder uses 2,827 MB of storage with 1,41 8ms on average inference time and 2,985 ms on average comparison time. Despite the success, these neural networks depend on high-powered environments, such as the millions or billions of parameters they possess, followed by the high computational resources, such as GPUs and extensive data needed to train the networks, which limits the deployment of these models on low-powered or edge devices.

\begin{figure}
    \centering
    \resizebox{.45\textwidth}{!}{
        \includegraphics{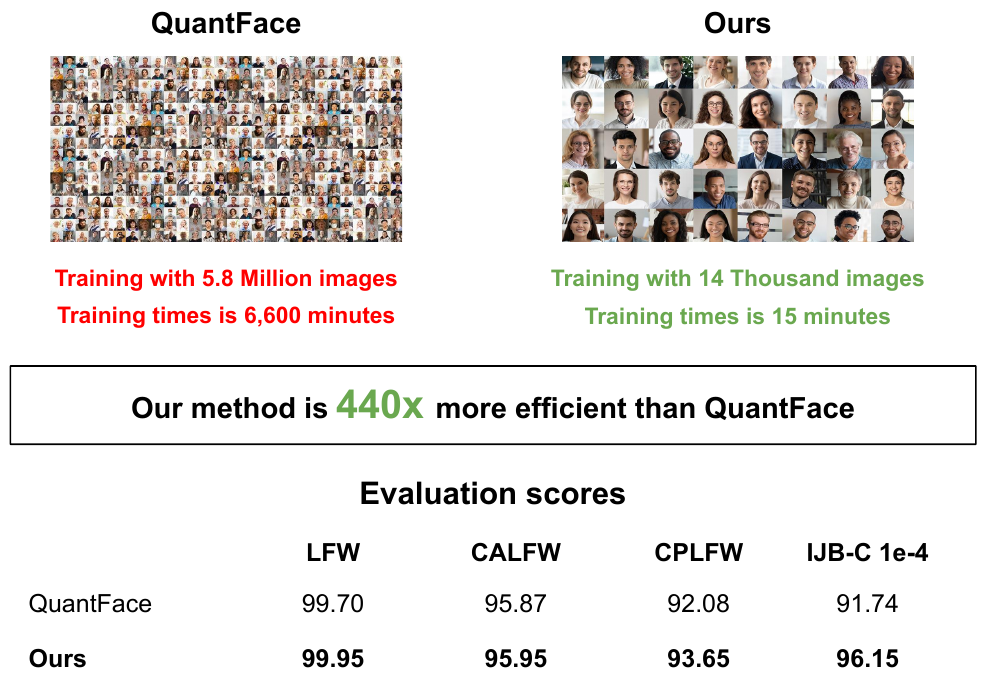}
    }
\caption{The illustration of our proposed solution compared to QuantFace~\cite{BoutrosICPR2022}. Our method efficiently reduced the training time by 440$\times$ while still achieving state-of-the-art performance. \textbf{Bold} text represents the best score.}
    \label{fig:compression-distribution}
    \vspace{-0.5cm}
\end{figure}

\begin{figure*}
    \centering
    \resizebox{1.\textwidth}{!}{
        \includegraphics{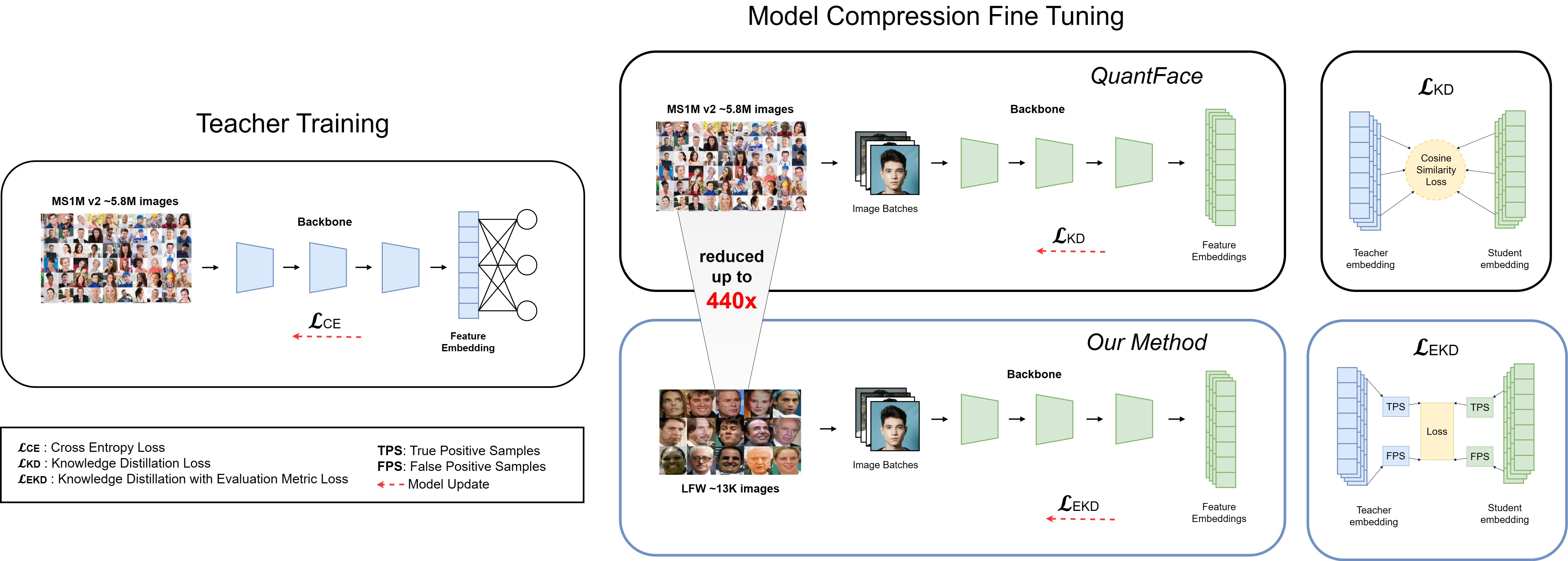}
    }
    \caption{An overview of our proposed pipeline. The compression of the teacher network is executed through either quantization or pruning techniques. After the model compression step, the fine-tuning process is initiated under a knowledge distillation paradigm, with information distilled from the final feature embedding layer.}
    \label{fig:compression-pipeline}
\end{figure*}

Within the context of model compression for FR, Boutros \textit{et al.} introduced QuantFace~\cite{BoutrosICPR2022} as a compression method for FR models, involving the quantization of the bit-width of network weights and activations. However, QuantFace requires 5.8M images, such as the MS1MV2 dataset, and half a million images of synthetic data, which makes it impractical to employ this in real-world scenarios. Additionally, large datasets require significant time and resources, even synthetic datasets \cite{BoutrosICPR2022}, often presenting practical challenges. This method is designed to address and overcome such constraints, leading to the question: \textit{is an extensive volume of training data truly necessary?} Moreover, there is a lack of extensive analysis in this domain regarding the superiority of one model compression technique over another and the corresponding performance trade-offs.

In this paper, we focus on efficient training for a quantized face recognition (FR) model, as illustrated in Figure~\ref{fig:compression-distribution}. Our approach revolves around a relatively small dataset created using the LFW dataset \cite{Huang2007LFW}, which consists of 13,233 images (round up to 14,000 images). Notably, this dataset is 440$\times$ smaller than the extensive MS1MV2 dataset used in QuantFace. In addition, we delve into an alternative training approach involving the Evaluation-oriented Knowledge Distillation (EKD)~\cite{HuangCVPR2022} to align with model quantization. Our method significantly improves the training efficiency of QuantFace by reducing the training time from the original 6,600 minutes per epoch to just 15 minutes per epoch, all while achieving state-of-the-art results within a single epoch. 
In the analysis section, we extensively explain our proposed solution and apply it to other compression methods, such as model pruning. 

%% file: sections_workshop/section2_related_work.tex
\section{Related Work}

\subsection{Face Recognition}
\label{sec:finetune}
Being a crucial biometric technique extensively employed in various industries \cite{WangNeurocomputing2021}, it is essential that the FR model deployed for achieving high performance is usable for critical applications such as security systems, biometric authentication, and healthcare. To achieve this goal, researchers have primarily focused on optimizing its performance in three principal domains: loss functions, network architectures, and datasets. 

The optimization of loss functions have been a primary focus in research, as they directly impact the performance potential of the model. This focus began with the development of angular/cosine margin loss \cite{DengCVPR2019} and its various adaptations. Another significant factor is the development of diverse network architectures, which have shown a trend of improving performance but often at the cost of increased parameters, making them unsuitable for implementation on edge devices \cite{WangNeurocomputing2021}. In addition to these factors, the training dataset also plays a vital role in the efficacy of face recognition. Datasets with over 1 million images and 100,000 identities, such as MS-Celeb-1M \cite{YandongECCV2016}, have become performance baselines.

\subsection{Model Compression}
\label{sec:related_work_compression}

Deep neural networks, especially state-of-the-art models such as deep convolutional neural networks (CNN) or transformers, often exhibit a substantial parameter count and demand extensive memory storage. This characteristic renders them impractical for training and deployment in a resource-constrained environment. Hence, this is where model compression emerges as a critical strategy, aiming to reduce the memory footprint and storage requirements, thereby enabling an efficient training and deployment process with minimal trade-offs in constrained scenarios. Various techniques have been introduced for model compression, including network quantization \cite{KrishnamoorthiArXiv2018, CaiCVPR2020, WuArXiv2020, NagelArXiv2021, BoutrosICPR2022}, parameter pruning \cite{ZhuICLR2018, HassibiICNN1993, DuggalICBD2021, FrankleICLR2019}, knowledge distillation \cite{HintonArXiv2015, HuangCVPR2022}, low-rank matrix factorization \cite{KumarArXiv2016, ZhangRTIP2023}, and designing compact neural network architectures \cite{IandolaArXiv2016, YanICCVW2019, ChenCCBR2018, MartinezICCVW2019}. In this study, we primarily focus on network quantization, and we employ knowledge distillation as the fine-tuning scheme. Moreover, we conduct a comprehensive experiment of pruning to show the performance of our proposed method across diverse compression techniques.
\vspace{-0.3cm}
\paragraph{Model Quantization.} Model quantization aims to reduce data precision by transitioning from high-precision formats such as FP32 to lower precision representations (e.g., INT8, binary \cite{HubaraNeurIPS2016}, or ternary \cite{LiArXiv2016}). The quantization process is influenced by several factors, including the selection between symmetric and asymmetric quantization, which is determined upon the zero-point value. This choice may involve training the quantized model, known as 'Quantization-Aware Training,' or adopting the 'Post-Training Quantization' paradigm, which allows direct deployment of the quantized model without additional training.

\paragraph{Model Quantization in FR} 
A relevant study exploring the quantization of face recognition networks was presented by Boutros \textit{et al.}~\cite{BoutrosICPR2022}, referred to as QuantFace. The fundamental concept of QuantFace involves the utilization of a feature-based Knowledge Distillation (KD) loss for training the model, employing the MS1MV2 dataset~\cite{DengCVPR2019}.

Our approach differs from QuantFace by focusing on significantly smaller datasets, demonstrating the potential for accuracy recovery even in scenarios of small data availability during compression. Additionally, we employ a modified version of the Evaluation-based Knowledge Distillation loss (EKD)~\cite{HuangCVPR2022} as the constrained training to specific FR tasks.

%% file: sections_workshop/section3_methodology.tex
\vspace{0.2cm}
\section{Proposed Method}

Figure~\ref{fig:compression-pipeline} illustrates the overall pipeline; the pre-trained model is first compressed using the aforementioned compression technique. Subsequently, the model undergoes a fine-tuning process employing the proposed scheme, enabling the model to learn effectively from small data. 

\paragraph{Efficient Quantization with EKD loss}
The common Knowledge Distillation (KD) paradigm is centered on transferring knowledge from the teacher network, typically a larger network, to accelerate the training of the student network, often a smaller network. This approach avoids the necessity to do a full retraining of the student network. Commonly, KD employs a specific loss function which is defined as follows:
\begin{equation}
    \mathcal{L}_{KD} = ||s(T) - t_k(T) - (s(S) - t_k(S))|| 
    \label{eq:kd_loss}
\end{equation}

In the equation presented, the variables are denoted as follows: $T$ represents the teacher network, $S$ represents the student network, $s$ represents the similarity score, and $t_k$ represents the threshold score within the range $k$. Notably, the algorithm we opt to use in our method differs from the conventional KD approach, known as Evaluation-oriented Knowledge Distillation (EKD)~\cite{HuangCVPR2022}. Specifically, EKD differentiates itself from the general KD by utilizing an evaluation-based metric as the primary concept of its loss function.

In the case of FR, the evaluation metrics used are True Positive Rate (TPR) and False Positive Rate (FPR), which can be seen in the following equation:
\begin{equation}
    FPR(t) = \frac{1}{M} \sum^{M}_{i=1}(v_i > t), \;
    TPR(t) = \frac{1}{M} \sum^{M}_{i=1}(u_i > t)
    \label{eq:fpr_trp}
\end{equation}
Where $v$ and $u$ are the negative pairs and positive pairs respectively and $t$ is the threshold. In this case, we use multiple thresholds with $k$ numbers ($t_k$) that correspond to the chosen range of FPR. The FPR range is set to [$1e-1, 1e-6$] following the standard face benchmarks. 
The positive $pos$ and negative $neg$ pairs are computed in one mini-batch during training by adapting the cosine similarity. The final EKD loss function is formulated as follows:
\begin{equation}
    \mathcal{L}_{ekd} = \lambda_{1}\mathcal{L}_{pos} + \lambda_{2}\mathcal{L}_{neg} + \mathcal{L}_{Arcface}
    \label{eq:ekd_ori}
\end{equation}
where $\lambda_{1}$ and $\lambda_{2}$ are weight parameters. For quantization, we modify Eq.~\ref{eq:ekd_ori} by excluding the classification loss $\mathcal{L}_{Arcface}$. The modified equation is formulated as follows:
\begin{equation}
    \mathcal{L}_{EKD} = \lambda_{1}\mathcal{L}_{pos} + \lambda_{2}\mathcal{L}_{neg}.
    \label{eq:ekd_quant}
\end{equation}
This modification is motivated by real-world scenarios where training lacks class labels. The removal of class labels not only addresses the absence of such labels but also proves advantageous in reducing training overheads, resulting in more efficient training.

%% file: sections_workshop/section4_experiment_details.tex
\section{Experiments}

\subsection{Dataset}
\paragraph{Prior Work Training Set.}
For training the teacher model, prior works use Glint360K Dataset \cite{AnICCVW2021} which includes 17M images with 360K identities. For training the compressed model, QuantFace uses two types of datasets for fine-tuning the network. In the first approach, they undertake the fine-tuning and calibration of the quantized network using the real image training data, specifically the MS1MV2 Dataset with 5.8M images of 85K identities. In the second approach, a set of 500K randomly generated synthetic face images are generated using StyleGAN2-ADA~\cite{KarrasCVPR2020}. 

\paragraph{Our Training Set.} 
We simulate the small data condition with the LFW~\cite{Huang2007LFW} training dataset, which consists of a mere 13,233 images distributed across 5,749 identities to train the compressed model. 

\begin{table*}
\centering
    \resizebox{1.\textwidth}{!}{
        \ra{1.3}
        \begin{tabular}{l l l c ccccc cc}
        \toprule\hline
        \multirow{2}{*}{Architecture} & \multirow{2}{*}{Params} & \multirow{2}{*}{Model Type} & Model Size & Total Training & LFW & CALFW & SLLFW & CPLFW & \multicolumn{2}{c}{IJB-C (TPR@FPR)} \\
        \cline{6-11}
        & & & (MB) & Images(M) & \multicolumn{4}{c}{Accuracy (\%)} & $1e^{-5}$ (\%) & $1e^{-4}$ (\%) \\
        \hline

        \multirow{5}{*}{R18} & 24.03M & \cellcolor{lightgray}{ } & \cellcolor{lightgray}{96.22} & \cellcolor{lightgray}{17.091M} & \cellcolor{lightgray}{99.88} & \cellcolor{lightgray}{95.75} & \cellcolor{lightgray}{99.53} & \cellcolor{lightgray}{90.80} & \cellcolor{lightgray}{93.16} & \cellcolor{lightgray}{95.33} \\
        & 24.03M & RealQuantFace~\cite{BoutrosICPR2022} & 18.10 & 5.811M & 99.52 & 95.58 & - & 88.37 & - & 93.03 \\
        & 24.03M & SynQuantFace & 18.10 & 0.500M & 99.55 & 95.32 & - & 89.05 & - & 90.38 \\
        & 24.03M & OursQuantized (w6a6) & 18.10 & 0.013M & \cellcolor{Goldenrod}{99.86} & 95.31 & \cellcolor{Goldenrod}{99.45} & 89.85 & 90.46 & 93.79 \\
        &  17.39 M  & OursPruned (sp02) & 61.39 & 0.013M & 99.11 & 91.94 &  80.73 & 80.63 & 54.19 & 76.66 \\
        \hline

        \multirow{3}{*}{R34} & 34.14 M & \cellcolor{lightgray}{ } & \cellcolor{lightgray}{130.44} & \cellcolor{lightgray}{17.091M} & \cellcolor{lightgray}{99.95} & \cellcolor{lightgray}{96.00} & \cellcolor{lightgray}{99.76} & \cellcolor{lightgray}{93.33} & \cellcolor{lightgray}{95.16} & \cellcolor{lightgray}{96.56} \\
        & 34.14 M & OursQuantized (w6a6) & 24.45 & 0.013M & \cellcolor{Apricot}{99.95} & \cellcolor{Apricot}{95.96} & \cellcolor{Apricot}{99.69} & \cellcolor{Goldenrod}{92.35} & \cellcolor{Goldenrod}{93.81} & \cellcolor{Goldenrod}{95.56} \\
        & 23.83 M & OursPruned (sp02) & 87.22 & 0.013M & 99.58 & 93.40 & 84.43 & 83.65 & 81.05 & 89.11 \\
        \hline

        \multirow{5}{*}{R50} & 43.59 M & \cellcolor{lightgray}{ } & \cellcolor{lightgray}{174.68} & \cellcolor{lightgray}{17.091M} & \cellcolor{lightgray}{99.95} & \cellcolor{lightgray}{96.06} & \cellcolor{lightgray}{99.76} & \cellcolor{lightgray}{94.40} & \cellcolor{lightgray}{95.61} & \cellcolor{lightgray}{96.97} \\
        
        & 43.59 M& RealQuantFace~\cite{BoutrosICPR2022} & 32.77 & 5.811M & 99.70 & 95.87 & - & 92.08 & - & 91.74 \\
        & 43.59 M& SynQuantFace~\cite{BoutrosICPR2022} & 32.77 & 0.500M & 99.68 & 95.70 & - & 90.38 & - & 90.72 \\
        & 43.59 M& OursQuantized (w6a6) & 32.77 & 0.013M & \cellcolor{Apricot}{99.95} & \cellcolor{Goldenrod}{95.95} & \cellcolor{Apricot}{99.69} & \cellcolor{Apricot}{93.65} & \cellcolor{Apricot}{94.13} & \cellcolor{Apricot}{96.15} \\
        & 29.83 M & OursPruned (sp02) & 111.32 & 0.013M & 99.73 & 94.08 & 86.98 & 85.43 & 83.24 & 91.08 \\

        \hline\bottomrule
        \ra{1.3}
        \end{tabular}
    } 
    \vspace{-0.5cm}
    \caption{Accuracy on LFW, CALFW, SLLFW, CPLFW, and IJB-C (TPR@FPR $1e^{-4}$ and $1e^{-5}$). Results are reported for ResNet18, ResNet34, and ResNet50 architectures from the baseline, quantized (weight to 6-bit and activation to 6-bit), and pruned (channel sparsity to 20\%) network settings. The box with \textcolor{Apricot}{Apricot} and \textcolor{Goldenrod}{Yellow} colors represent the best and second best scores respectively. The box with \textcolor{lightgray}{gray} represents the teacher}
    \label{table:table_result_all}
    \vspace{-0.3cm}
\end{table*}

\paragraph{Test Set.} 
The validation dataset reported will be the standard dataset used for benchmarking FR models, including a range of datasets designed to assess various challenges and scenarios in FR. The dataset covers Labeled Faces in the Wild (LFW)~\cite{Huang2007LFW}, along with its extension datasets, namely Cross-Age LFW (CALFW)~\cite{ZhengArXiv2017}, Cross-Pose LFW (CPLFW)~\cite{Zheng2018CPLFW}, and Similar-Looking LFW (SLLFW)~\cite{DengPR2017}. We also include using Iarpa Janus Benchmark-C (IJB-C)~\cite{MazeICB2018} dataset consisting of approximately 3,500 identities and a total of around 469,375 images, allowing extensive testing and evaluation of FR models for wide scenarios and conditions. 

\subsection{Experimental setting}
\paragraph{Model Architecture.} We utilize the ResNet18 (R18), ResNet34 (R34), and ResNet50 (R50) models, which are accessible from InsightFace \cite{DengPR2017}, as our teacher models. In the instance of QuantFace, we utilize their pre-trained weights for the compressed models directly obtained from their original implementations on ResNet18, ResNet50, and ResNet100. This enables us to compare performance and behaviors across various architectures, as compressing different architectures may lead to divergent results, as noted in \cite{BoutrosICPR2022}.

\vspace{-5pt}
\paragraph{Method Comparison. }For comparison purposes, we categorize QuantFace results into two models: 'RealQuantFace,' trained with 5.8M MS1MV2 dataset, and 'SynQuantFace,' trained with synthetic face images. This allows us to compare our experiments with those presented in \cite{BoutrosICPR2022}, encompassing training on real and synthetic datasets.

\paragraph{Training.} 
Our teacher networks are sourced from InsightFace~\cite{DengPR2017}, ensuring easy accessibility and minimizing overall training time. The student network, representing the compressed model, is trained for a single epoch. To maintain consistency and fairness, we apply identical settings for both methods. These settings include a batch size of 64, a learning rate of 0.001, a margin of 0.4, and a scale of 64. Input images are resized to dimensions of 112 x 112, with an embedding size set to 512. All experiments are conducted using the PyTorch~\cite{PaszkeNeurIPS2019} framework.
Our hardware setup comprises an Intel(R) i7-7700 CPU running at 3.60 GHz, 32 GB of RAM, and an NVIDIA TITAN Xp GPU.

\paragraph{Evaluation Metrics}
We evaluate each dataset, considering the metrics used in related work. We use accuracy as the evaluation metric for all variations of the LFW dataset; specifically, we used LFW, CALFW, SLLFW, and CPLFW, following the approach in~\cite{Huang2007LFW}. For the IJB-C dataset, we follow the 1:1 verification protocol mentioned in ArcFace~\cite{DengCVPR2019}, where we calculate the average of image features as the corresponding template representation for $1e^{-4}$ and $1e^{-5}$ thresholds.

%% file: sections_workshop/section5_results.tex
\section{Results \& Analysis}

\begin{table*}
\centering
    \resizebox{1.\textwidth}{!}{
        \ra{1.3}
        \begin{tabular}{l l l cccccc cc}
        \toprule\hline
        \multirow{2}{*}{Architecture} & \multirow{2}{*}{Params} & \multirow{2}{*}{Model Type} & \multirow{2}{*}{Model Size} & Total training & LFW & CALFW & SLLFW & CPLFW & \multicolumn{2}{c}{IJB-C (TPR@FPR)} \\
        \cline{6-11}
        & & & (MB) & Images(M) & \multicolumn{4}{c}{Accuracy (\%)} & $1e^{-5}$ (\%) & $1e^{-4}$ (\%) \\
        \hline

        \multirow{2}{*}{R34} & \multirow{2}{*}{34.14 M} &  OursQuantized (w8a8) & 32.61 & 0.013M & \cellcolor{green!30}{99.95} & 95.93 & \cellcolor{green!30}{99.76} & \cellcolor{green!30}{93.23} & \cellcolor{green!30}{94.78} & \cellcolor{green!30}{96.38} \\
        && OursQuantized (w6a6) & 24.45 &  0.013M & \cellcolor{Apricot}{99.95} & \cellcolor{Apricot}95.96 & \cellcolor{Apricot}{99.69} & \cellcolor{Apricot}92.35 & \cellcolor{Apricot}{93.81} & \cellcolor{Apricot}95.56 \\
        
        \hline

        \multirow{4}{*}{R50} & \multirow{4}{*}{43.59 M} &  RealQuantFace (w8a8)~\cite{BoutrosICPR2022} & 43.67 & 5.811M & 99.78 & 96.00 & - & 92.17 & - & 95.66 \\
        &&  SynQuantFace(w8a8)~\cite{BoutrosICPR2022} &43.67 & 0.500M & 99.78 & 95.87 & - & 92.08 & - & 93.67 \\
        && RealQuantFace (w6a6)~\cite{BoutrosICPR2022} & 32.77 & 5.811M & \cellcolor{Goldenrod}99.70 & \cellcolor{Goldenrod} 95.87 & - & \cellcolor{Goldenrod}92.08 & - & \cellcolor{Goldenrod}91.74 \\
        && SynQuantFace(w6a6)~\cite{BoutrosICPR2022} & 32.77& 0.500M & 99.68 & 95.70 & - & 90.38 & - & 90.72 \\
        
        \hline
        
        \multirow{4}{*}{R100} & \multirow{4}{*}{65.2 M} & {RealQuantFace (w8a8)~\cite{BoutrosICPR2022}} & 65.31 & 5.811M & \cellcolor{purple!25}{99.80} & \cellcolor{green!30}{96.05} & - & \cellcolor{purple!25}{92.92} & - & \cellcolor{green!30}{96.38} \\
        &&  SynQuantFace(w8a8)~\cite{BoutrosICPR2022} & 65.31& 0.500M & \cellcolor{purple!25}{99.80} & \cellcolor{purple!25}{96.02} & - & 92.90 & - & \cellcolor{purple!25} 96.09 \\
        && RealQuantFace(w6a6)~\cite{BoutrosICPR2022} & 49.01 & 5.811M & 99.55 & 95.42 & - & 85.63 & - & 85.80 \\
        && SynQuantFace(w6a6)~\cite{BoutrosICPR2022} & 49.01& 0.500M & 99.45 & 95.58 & - & 86.60 & - & 87.00 \\
        \hline\bottomrule
        \ra{1.3}
        \end{tabular}
    } 
    \vspace{-0.5cm}
    \caption{
        Accuracy comparison between our method and the QuantFace method. results are reported on ResNet34 for our method against ResNet50 and ResNet100 for QuantFace. Two quantization methods, namely w6a6 (weight to 6-bit and activation to 6-bit) and w8a8 (weight to 8-bit and activation to 8-bit), are considered for each experiment reported in this table. The box with \textcolor{Apricot}{Apricot} and \textcolor{Goldenrod}{Yellow} colors represent the best and second best scores for w6a6 respectively. The box with \textcolor{purple!25}{Pink} and \textcolor{green!30}{Lime} colors represent the best and second best scores for w8a8 respectively.
    }
    \label{table:abl_ekd_loss}
    \vspace{-0.4cm}
\end{table*}

\paragraph{Comparison to QuantFace. } Table~\ref{table:table_result_all} illustrates a comparative analysis between our proposed solution and QuantFace. Our quantization approach achieves state-of-the-art performance across all standard evaluation datasets, even with only 0.132M data used. Specifically in R50, our quantization achieves 99.95, 99.69, 93.65, 94.13, and 96.15 on LFW, SLLFW, CPLFW, and IJB-C datasets respectively. Compared to RealQuantFace and SynQuantFace, our solution outperforms with a considerable gap of 4.41 and 5.43 on the IJB-C dataset $1e^{-4}$. It is noteworthy that our employed dataset is 440$\times$ smaller than that of RealQuantFace and 38.46$\times$ smaller than SynQuantFace. 

We analyzed our proposed solution implemented on the R34 architecture, comparing its performance against RealQuantFace and SynQuantFace on the R50 architecture. Our proposed solution consistently outperforms them, demonstrating a substantial margin of 3.82 and 4.84 on the IJB-C dataset $1e^{-4}$. Notably, our model is 7.32 MB smaller, with 9.45M fewer parameters compared to QuantFace on R50.

\begin{table}
\centering
    \resizebox{0.5\textwidth}{!}{
        \ra{1.3}
        \begin{tabular}{l l l  c cc}
        \toprule\hline
        \multirow{2}{*}{Architecture} & \multirow{2}{*}{Params} & \multirow{2}{*}{Model Type} & Total Training  & \multicolumn{2}{c}{IJB-C (TPR@FPR)} \\
        \cline{5-6}
        & & &  Images(M) & $1e^{-5}$ (\%) & $1e^{-4}$ (\%) \\
        \hline

        \multirow{2}{*}{R18}  
        & 24.03M & OursQuantized (w6a6)  & 0.013M & 90.46 & 93.79 \\
        & 24.03M & OursQuantized with CE (w6a6)  & 0.013M & \cellcolor{Apricot}{90.56} & \cellcolor{Apricot}{93.81} \\
        \hline

        \multirow{2}{*}{R34}
        & 34.14 M & OursQuantized (w6a6) & 0.013M & \cellcolor{Apricot}{93.81} & 95.56 \\
        & 34.14 M  M & OursQuantized (w6a6) with CE  & 0.013M & 92.91 &  \cellcolor{Apricot}{95.57} \\
        \hline

        \multirow{2}{*}{R50} 
        & 43.59 M& OursQuantized (w6a6)  & 0.013M &  \cellcolor{Apricot}{94.13} & \cellcolor{Apricot}{96.15} \\
        & 43.59 M & OursQuantized (w6a6) with CE  & 0.013M &  94.09 & 96.06 \\

        \hline\bottomrule
        \ra{1.3}
        \end{tabular}
    } 
    \vspace{-0.5cm}
    \caption{Accuracy on IJB-C (TPR@FPR $1e^{-4}$ and $1e^{-5}$). Results are reported for ResNet18, ResNet34, and ResNet50 architectures from the quantized (weight to 6-bit and activation to 6-bit) with classifier head and without classifier head marked by 'with CE'. The box with \textcolor{Apricot}{Apricot} colors represent the best scores for each network architecture. }
    \label{table:table_abl_ce}
    \vspace{-0.3cm}
\end{table}

\begin{table*}
    \centering
    \resizebox{1.\textwidth}{!}{
        \begin{tabular}{c cc cccc cc}
            \toprule\hline
            \multirow{2}{*}{Architecture} & Percentage of & Total Training & LFW & CALFW & SLLFW & CPLFW & \multicolumn{2}{c}{IJB-C (TPR@FPR)} \\
            \cline{4-9}
            
            & Used Images(\%) & Images & \multicolumn{4}{c}{Accuracy (\%)} & $1e^{-5}$ (\%) & $1e^{-4}$ (\%) \\
            \hline

            & 100\% & 13,233 & \cellcolor{Apricot}{99.95} & \cellcolor{Goldenrod}{95.96} & 99.69 & \cellcolor{Goldenrod}{92.35} & \cellcolor{Apricot}{93.81} & 95.56 \\
            \multirow{2}{*}{R34 (w6a6)} & 75\% & 9,924 & \cellcolor{Apricot}{99.95}  & \cellcolor{Apricot}{95.98} & 99.68 & \cellcolor{Apricot}{92.36} & 92.37 & 95.35 \\ 
            & 50\% & 6,616 & \cellcolor{Apricot}{99.95} & 95.91 & \cellcolor{Apricot}{99.73} & 92.11 & \cellcolor{Goldenrod}{93.02} & \cellcolor{Apricot}{95.79} \\
            & 25\% & 3,308 & \cellcolor{Goldenrod}{99.93} & 95.86 & \cellcolor{Goldenrod}{99.70} & 92.01 & 91.91 & \cellcolor{Goldenrod}{95.58} \\
            \hline

            \hline\bottomrule
            
        \end{tabular}
    }
    \caption{Accuracy on limited data count used, specifically on the LFW dataset. The box with \textcolor{Apricot}{Apricot} and \textcolor{Goldenrod}{Yellow} colors represent the best and second best scores respectively.}
    \vspace{-0.4cm}
    \label{table:table_result_id}
\end{table*}

\paragraph{Evaluation-Based Metric Loss for Quantization FR Model.}
We further analyze the effectiveness of Evaluation-based metric loss for quantization in the context of FR models as shown in Table~\ref{table:abl_ekd_loss}. We reduce the original precision to w8a8 (8-bit weights and 8-bit activations) and w6a6 (6-bit weights and 6-bit activations), following the QuantFace settings, on the ResNet34 (R34) model. We compare our R34 model with QuantFace trained on ResNet50 (R50) and ResNet100 (R100) for both synthetic and real data.

Our model achieves a performance level comparable to that of R100 w8a8, reaching 96.38\% accuracy as reported in IJB-C $1e^{-4}$, while employing significantly fewer parameters, approximately 52\% of the parameters in R100 and 2$\times$ smaller model size. 

In the w6a6 settings, it is evident that our training method can achieve better overall results with 95.56\%, as reported in IJB-C $1e^{-4}$, outperforming both R50 and R100. This is because evaluation-based metric loss was used as the objective function, which constrained the compressed model to perform a specific task.   

\vspace{-0.25cm}

\paragraph{Removing Class-label Effect.}

Table ~\ref{table:table_abl_ce} provides details on two scenarios in which each model is trained. Specifically, we employ two identical training schemes with the only difference being the inclusion of the classifier head. The results indicate that there is not a significant margin of accuracy difference between training with and without the Cross-Entropy (CE) loss. overall this pattern is observed through all the different architecture, for instance, in the R18 network, difference is only 0.10\% and 0.02\%, as reported in IJB-C $1e^{-4}$ and $1e^{-5}$ respectively, and in the R50 network, the difference is only 0.04\% and 0.09\%, as reported in IJB-C $1e^{-4}$ and $1e^{-5}$ respectively. This suggests that there is not a significant difference, leading us to argue that we can further reduce extra overheads and parameters trained during the training period by removing the classifier head entirely, as it does not have a substantial impact when training with limited data.

\paragraph{Quantization with Small Data.}
In Table~\ref{table:table_result_id}, we investigate the impact of employing different quantities of images for training a quantized network.  Additionally, employing only 25\% of the LFW dataset, approximately 3,300 images, leads to a notable improvement in model performance, achieving an accuracy of 95.58\% on the IJB-C $1e^{-4}$ dataset.

Beyond this point, the impact on model performance became increasingly marginal as the proportion of LFW dataset images used increased. Using 50\% of the LFW dataset (approximately 6,600 images) elevated the accuracy to 95.79\% in IJB-C $1e^{-4}$. Expanding the dataset to 75\% (around 9,900 images) resulted in a slight decrease in performance to 95.35\% in IJB-C $1e^{-4}$. Training with the entire LFW dataset, approximately 13,200 images, achieved an accuracy of 95.56\% in the IJB-C $1e^{-4}$ benchmark. In theory, quantization is aimed at reducing the precision of the model, making it faster than a full-precision model. This principle aligns with the theory of quantization, which emphasizes the model's capacity to adapt and align itself with the specific task \cite{ChengArXiv2017}. This observation underscores the potential of quantization to yield benefits, even when employed with small data. 

\paragraph{Small Data Utilization for Different Model Compression.}
We simulate with ResNet34 (R34) as the target compression model. The objective is to compress R34 with a 20\% pruned channel, denoted by 'Pruned (sp02)' using DepGraph~\cite{FangArXiv2023} with the aim of achieving a successful reduction in the number of parameters from ResNet34 (R34) to ResNet18 (R18).

In Table~\ref{table:table_result_all}, we observe that the results of our pruned model with small data are still far from R18, with a 6.22\% gap on the IJB-C dataset. This could be attributed to the fact that pruning directly removes network parameters, and a small dataset may not provide sufficient information to recover the lost knowledge \cite{FrankleICLR2019}. This may occur because the pruning is actually changing the whole network, therefore using a small dataset is insufficient. 
In contrast, as quantization solely reduces data type precision without changing the network architecture, it is sufficient to just calibrate the model with small datasets~\cite{ChengArXiv2017}. 

\begin{table}
\centering
    \resizebox{0.5\textwidth}{!}{
        \ra{1.3}
        \begin{tabular}{l l l c c c c}
        \toprule\hline
        \multirow{2}{*}{Methods} & \multirow{2}{*}{Params} & \multirow{2}{*}{Model Type} & \multirow{2}{*}{Model Size} & Total training & \multicolumn{2}{c}{IJB-C (TPR@FPR)} \\
        \cline{6-7}
        & & & (MB) & Images(M) & $1e^{-5}$ (\%) & $1e^{-4}$ (\%) \\
        \hline
        Teacher(EKD) & 43.6M & R50 & 174.68 & 5.811M & 95.61 & 92.66 \\
        Teacher(quant) & 24.03M & R18 & 96.22 & 96.22M & 93.16 & 95.33 \\
        \hline
        EKD~\cite{HuangCVPR2022} & 24.03M & R18 & 96.22M & 5.811M & 88.84 & 92.74\\ 
        \hline
        QuantFace(w6a6)~\cite{BoutrosICPR2022} & 24.03M & R18 & 18.10M & 5.811M & - & 93.03\\ 
        OursQuantized(w6a6) & 24.03M & R18 & 18.10M  & 0.132M & 90.46 & 93.79\\ 
        \hline
        
        \end{tabular}
    } 
    \caption{Accuracy of IJB-C (TPR@FPR $1e^{-4}$ and $1e^{-5}$). Row 1 and 2 represent the teacher model. Row 3 is the Knowledge distillation-based model compression. Row 4 and 5 are the quantization-based model compression.}
    \label{table:abl_kd}
    \vspace{-0.4cm}
\end{table}

\paragraph{Analysis for Model Compression in FR}
We perform an extensive analysis of various compression methods, as presented in Table~\ref{table:abl_kd}. In the context of FR, Knowledge Distillation (KD) is a common approach as the compression technique. Table~\ref{table:abl_kd} illustrates an example of the distillation result using EKD from R50 to R18. In contrast, the quantization-based approach reduces data precision within the same model architecture (e.g., from R18 to R18). Our experiments reveal that KD requires training with substantial data and relies on classification loss.

Table~\ref{table:abl_kd} demonstrates the performance of the KD-based approach, achieving 92.74\% and 88.84\% accuracy on the IJB-C dataset at $1e^{-4}$ and $1e^{-5}$, respectively. On the other hand, quantization requires only a small dataset to achieve state-of-the-art performance. It is important to note that although quantization aims to reduce the bit-width of weights, the overall parameter count may not necessarily decrease. In addition, the post-processing is required to compute the model size on the quantized model. 

In practical scenarios, quantization is often recommended due to its ease of implementation and practical usefulness. It significantly reduces inference time, making it a valuable choice for real-world applications.

%% file: sections_workshop/section6_conclusion.tex
\vspace{-0.2cm}
\section{Conclusion}
Our study is centered on efficient training for quantized models in the domain of face recognition. Typically, for robust face recognition models, a substantial dataset is required, and this also holds true for calibrating a quantized model. Consequently, we questioned whether a massive dataset is an absolute necessity for this task. 

Our proposed training paradigm demonstrates that our method can achieve state-of-the-art results, even when compared to models with more parameters and using a considerably larger datasets. Additionally, we show that our approach significantly reduces training time, a critical consideration in the research field, which then facilitates the execution of more experimental investigations. This not only support constrained settings where accelerated training schemes are not feasible but also proves advantageous for fine-tuning a compressed model.

%% file: sections_workshop/section7_future_works.tex
\section{Discussion and Future Works}

In Table~\ref{table:abl_ekd_loss}, we observed a decrease in accuracy as we applied quantization to more extensive networks. For instance, in R50, the accuracy difference between w8a8 and w6a6 is 3.92\%, as reported in IJB-C $1e^{-4}$ when trained on real data. However, this difference becomes more pronounced as the model parameters increase. In R100, the accuracy gap between w8a8 and w6a6 widens to 10.58\%, as reported in IJB-C $1e^{-4}$ with the same training data. Such degradation observed in this approach restricts its practical applicability, and this prompts our interest to further investigate the challenges associated with quantizing larger models. Furthermore, there is potential for in-depth exploration of low-bit quantization (\textit{e.g.} 2-bits) in the context of deep facial recognition models through comprehensive experiments in the future.